\documentclass[journal,twoside,web]{ieeecolor}
\usepackage{generic}
\usepackage{cite}
\usepackage{amsmath,amssymb,amsfonts}
\usepackage{algorithmic}
\usepackage{graphicx}
\usepackage{algorithm,algorithmic}
\usepackage{hyperref}
\hypersetup{hidelinks}
\usepackage{textcomp}
\usepackage{ragged2e}
\usepackage{array}
\usepackage{subcaption}
\usepackage{tabularx}
\usepackage{booktabs}
\usepackage{multirow}
\usepackage{orcidlink}

\def\BibTeX{{\rm B\kern-.05em{\sc i\kern-.025em b}\kern-.08em
    T\kern-.1667em\lower.7ex\hbox{E}\kern-.125emX}}
\begin{document}
\title{Automated Medical Report Generation for ECG Data: Bridging Medical Text and Signal Processing with Deep Learning}

\author{Amnon Bleich\,\orcidlink{0009-0001-9475-4668}, 
Antje Linnemann\,\orcidlink{0009-0002-5644-5845}, 
Björn H. Diem\,\orcidlink{0009-0002-7032-8100}, and 
Tim~OF~Conrad\,\orcidlink{0000-0002-5590-5726}
\thanks{This work was supported by the German Ministry for Education and Research (BMBF) within the Berlin Institute for the Foundations of Learning and Data - BIFOLD (project grants 01IS18025A and 01IS18037I) and the Forschungscampus MODAL (project grant 3FO18501).}%
\thanks{Amnon Bleich and Tim~OF~Conrad are with Zuse Institut Berlin, Takustraße 7, 14195 Berlin, Germany (e-mail: bleich@zib.de; conrad@zib.de).}%
\thanks{Antje Linnemann and Björn H. Diem are with BIOTRONIK SE \& Co KG, Woermannkehre 1, 12359 Berlin, Germany (e-mail: antje.linnemann@biotronik.com; bjoern.diem@biotronik.com).}%
}

\maketitle

\begin{abstract}
Recent advances in deep learning and natural language generation have significantly improved image captioning, enabling automated, human-like descriptions for visual content. In this work, we apply these captioning techniques to generate clinician-like interpretations of ECG data. This study leverages existing ECG datasets accompanied by free-text reports authored by healthcare professionals (HCPs) as training data. These reports, while often inconsistent, provide a valuable foundation for automated learning. We introduce an encoder-decoder-based method that uses these reports to train models to generate detailed descriptions of ECG episodes. This represents a significant advancement in ECG analysis automation, with potential applications in zero-shot classification and automated clinical decision support.

The model is tested on various datasets, including both 1- and 12-lead ECGs. It significantly outperforms the state-of-the-art reference model by Qiu et al., achieving a METEOR score of 55.53\% compared to 24.51\% achieved by the reference model.
Furthermore, several key design choices are discussed, providing a comprehensive overview of current challenges and innovations in this domain.

The source codes for this research are publicly available in our Git repository\footnote{\raggedright\url{https://git.zib.de/ableich/ecg-comment-generation-public}}.
\end{abstract}


\section{Introduction}
\label{sec:introduction}
\IEEEPARstart{C}{ardiovascular} diseases remain a leading cause of morbidity and mortality worldwide, highlighting the importance of effective and timely diagnostic methods. Traditional ECG analysis, while indispensable, often requires manual interpretation by trained physicians, which can be time-consuming and subject to human error. At the same time, advances in AI-driven text generation algorithms present a novel opportunity to take advantage of existing ECG data sets for which physician-written free-text comments are available. These models have the potential to generate informative text about ECG episodes, which could offer decision support to clinicians during diagnostic tasks and thus reduce cognitive load on medical professionals and potentially improve the speed and accuracy of diagnosis.

Significant progress in encoder-decoder architectures, particularly in image captioning, has demonstrated their capacity for generating coherent descriptions \cite{vinyals2015show, xu2015show}. While these methods have succeeded in domains such as image processing, their application to medical data, specifically electrocardiography (ECG), remains underexplored \cite{pang2023survey}. One major challenge is the scarcity of purpose-built, high-quality labeled datasets for ECG analysis, as generating such datasets requires the involvement of highly trained individuals, making the process resource-intensive \cite{bleich2023enhancing}. However, a promising alternative lies in leveraging existing ECG records paired with free-text comments, which are often generated as a byproduct of routine clinical practices \cite{johnson2023mimic}. Although these free-text comments were not originally created for training machine learning models, they represent a rich source of data that can be adapted to train encoder-decoder architectures, capable of generating concise textual summaries of ECG episodes, as demonstrated in this work.

\begin{figure*}[!htpb]
    \centering
    \includegraphics[width=0.9\textwidth]{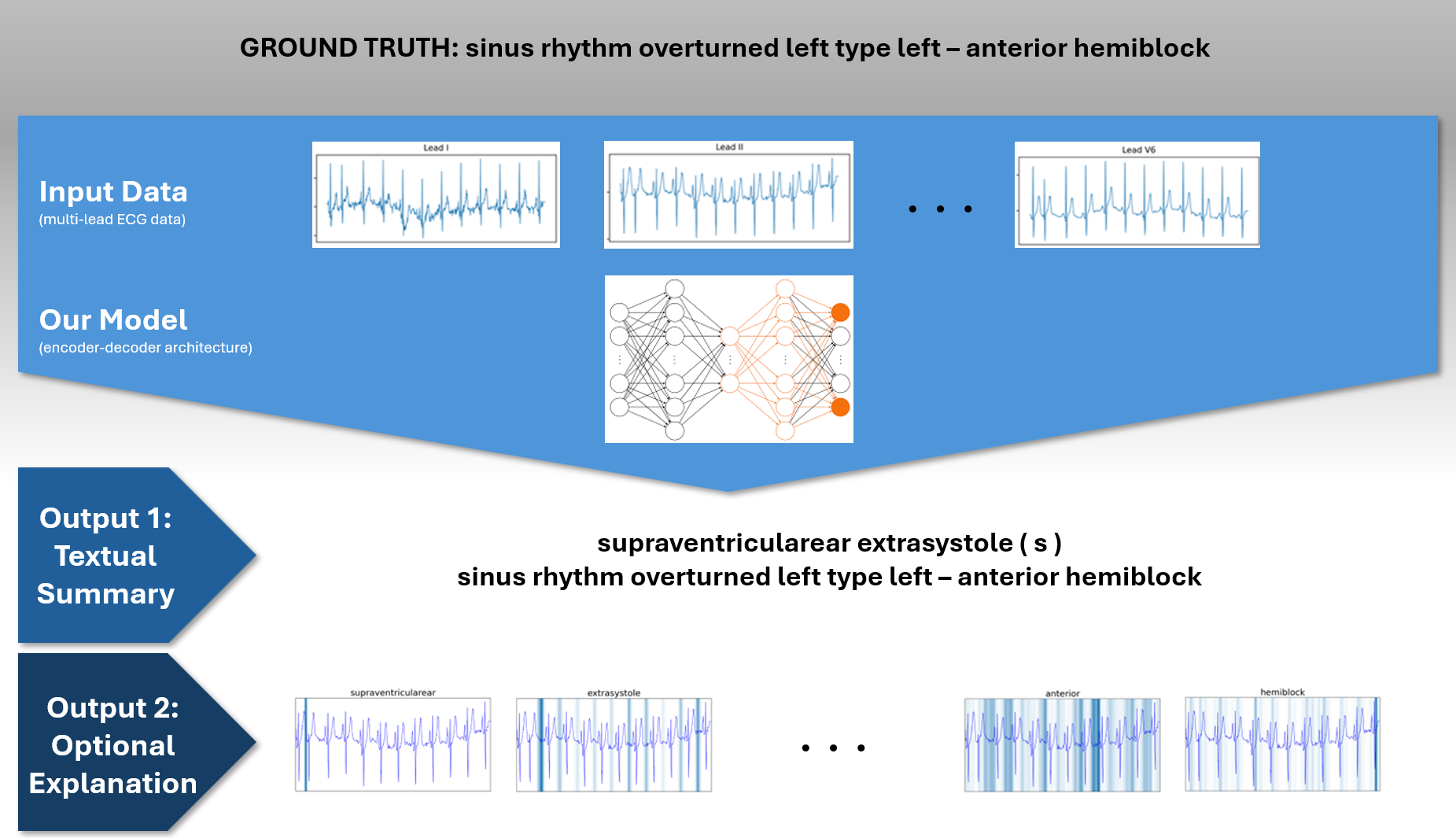}
    \caption{Overview of our proposed method for automated ECG textual summary generation. The input consists of multi-lead ECG data, from which our model generates a textual summary. Additionally, the method identifies and highlights supporting diagnostic features in the ECG signal (e.g., "extrasystole," "anterior," and "hemiblock") as part of the generated explanation, providing interpretability to support clinical decision-making. Note that the ground truth shown at the top in this example is not available to the model.}
        \label{fig:method_overview}
\end{figure*}

This study introduces an innovative approach to text generation designed for ECG analysis, validated on the publicly available PTB-XL dataset \cite{wagner2020ptb}. In addition to achieving state-of-the-art performance, a key contribution of this work is the establishment of a reproducible benchmark for future studies, as our work is, to the best of our knowledge, first of its kind to be tested on the official test subset of the PTB-XL dataset, ensuring both transparency and comparability. 
Furthermore, we present a case-study on a more challenging and larger dataset of single-lead, variate, and subcutaneous ECG (sECG) signals obtained from implantable cardiac monitors (ICMs) \cite{bleich2023enhancing}. Unlike purpose-built datasets, this dataset comprises labels and free-text reports generated as a byproduct of routine clinical and administrative processes. These reports, often created to fulfill procedural requirements, have not been cross-checked for accuracy, introducing additional noise and complexity that make the dataset more difficult to leverage for automated learning. This case study highlights the robustness and adaptability of our method to diverse and less curated real-world data sources.

The method introduced in this study employs an encoder-decoder architecture, utilizing a ResNet-based encoder \cite{he2016deep} paired with either an LSTM \cite{hochreiter1997long} or Transformer \cite{NIPS2017_3f5ee243} decoder.

In the following sections, we provide an overview of related work to contextualize this study within the broader scientific landscape and facilitate comparison with existing methods. We then present a comprehensive description of our proposed model, detailing its architecture, tokenization techniques, data preprocessing procedures, and the hyperparameter optimization and training strategies employed. This is followed by a description of the datasets used, along with the specific preprocessing steps applied. Next, we outline the experiments conducted to evaluate the performance of our model and analyze the results obtained. Finally, we discuss the key findings and suggest directions for future research.

\section{Related Work} \label{sec:related-work}
Various methods have been proposed in recent years for the automatic diagnosis of ECG data (e.g., \cite{moghadam2023automatic, denysyuk2023algorithms, wang2023hierarchical}) and other feature-based heart data (e.g., \cite{Ahmed2019}). Stracina et al. \cite{stracina2022golden} provide a comprehensive review of advancements and possibilities in ECG analysis since the first successful recording of the electrical activity of the human heart in 1887, including recent developments leveraging deep learning techniques.

A key takeaway from these reviews is that the majority of existing ECG analysis methods focus primarily on the classification of ECG episodes, such as identifying atrial fibrillation (AF) or other abnormal heart rhythms. In contrast, the field of automated report generation for ECG data remains largely unexplored. Meanwhile, significant advancements in the task of generating textual descriptions from visual data (commonly referred to as image captioning) have been achieved through the development of sophisticated machine learning models and language models in particular. Seminal works in this domain include works by Vinyals et al. \cite{vinyals2015show}, and subsequently Xu et al. \cite{xu2015show}, which introduced an encoder-decoder architecture utilizing ResNet \cite{he2016deep} as the encoder and a Long Short-Term Memory (LSTM) network \cite{hochreiter1997long} as the decoder to successfully generate descriptive text for images. This framework has since been extended to various domains, including medical signal processing. Pang et al. \cite{pang2023survey} conducted a survey of recent technologies in medical report generation, primarily addressing medical imaging, such as X-ray screenings, demonstrating the versatility and potential of such architectures. However this work offers limited insight into one-dimensional signals like ECG or EEG. This highlights a critical gap in the application of LLMs to medical data, emphasizing the need for further exploration in this area.

Given the substantial progress in automated data captioning and the relative scarcity of work in applying these techniques to ECG data, this area represents a significant research opportunity. Building on the foundational work of Vinyals et al. and Xu et al., this study explores the application of similar encoder-decoder architectures to the generation of free-text descriptions for ECG episodes, aiming to bridge this gap in the field.

One notable effort in this direction is the work by Qiu et al. \cite{qiu-etal-2023-transfer}, who investigated the potential of pre-trained large language models (LLMs) for ECG classification. They demonstrated that knowledge from natural language processing could be effectively transferred to ECG data, enabling the detection of cardiovascular diseases. Their approach involved fine-tuning pre-trained LLMs on the publicly available PTB-XL dataset \cite{wagner2020ptb}, achieving highly encouraging performance. However, their method faced certain limitations, such as random data splitting, which can lead to overfitting, as evident by the minimal performance gains after incorporating the ECG data itself into the model (refer to Section \ref{sec:sanity-check} for further details). Since their work was not tested on the PTB-XL official splits and no other studies have reported results on publicly available datasets, we used their method as a reference model for performance comparison of our model (refer to Section \ref{sec:baseline}).

Bartels et al. \cite{bartels2022learning} explored a method using a private dataset of ECG records with captions, partially machine encoded and corrected by healthcare professionals (HCPs), in addition to those generated by HCPs during routine medical practices. Their method used a pre-trained ECG classification ResNet (ECGNet \cite{van2020automatic}) with an additional classification layer to embed signals and feed the embedded signals into an LSTM or a Transformer. However, their approach did not include testing on publicly available datasets, which limits the generalizability of their model evaluations. Furthermore, they did not explore using non-pre-trained models or single-lead ECG signals, and their proposed model architecture lacked detailed explanation.

In conclusion, while caption generation for images has achieved satisfactory levels, the task remains complex for medical images and even more so for ECG data. Current advances in applying LLMs to ECG analysis are promising, but a proven approach to successfully evaluate public datasets with trustworthy evaluation metrics is still being investigated. Our work aims to address this gap by leveraging free-text data and exploring the potential of modern machine learning techniques in the challenging domain of ECG signal interpretation.

\section{A New Method for ECG Caption Generation: Architecture and Training Optimization}\label{sec:method}

This section introduces a novel approach for generating text descriptions of ECG data through encoder-decoder architectures, adapted from successful image captioning frameworks \cite{vinyals2015show, xu2015show}. The method leverages a ResNet-based encoder for embedding ECG signals and incorporates Transformer or LSTM decoders to produce descriptive, clinically relevant reports.

\begin{figure*}[!htpb]
    \centering
    \begin{subfigure}[t]{\textwidth} 
        \centering
        \includegraphics[width=\textwidth]{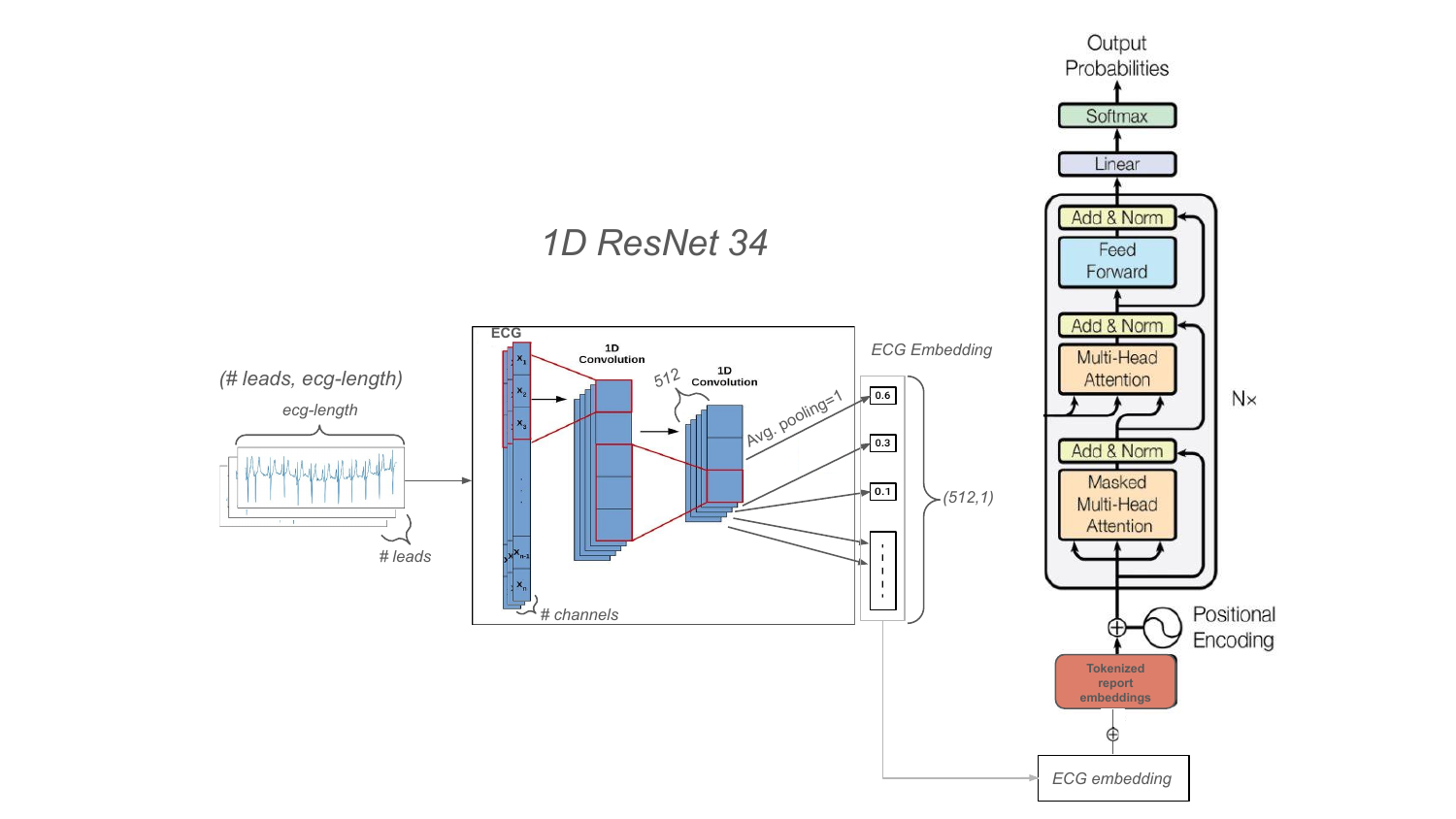}
        \caption{Our Transformer-based model architecture: The ECG signal is input into the 1D ResNet34, producing 512 channels, each of a user-defined size $x$ ($x=1$ in our best-performing model). The $x$ vectors are concatenated with the tokenized report embeddings and used as input to the Transformer model.}
    \label{fig:transformer_architecture}
    \end{subfigure}
    \begin{subfigure}[t]{\textwidth} 
        \centering
        \includegraphics[width=\textwidth]{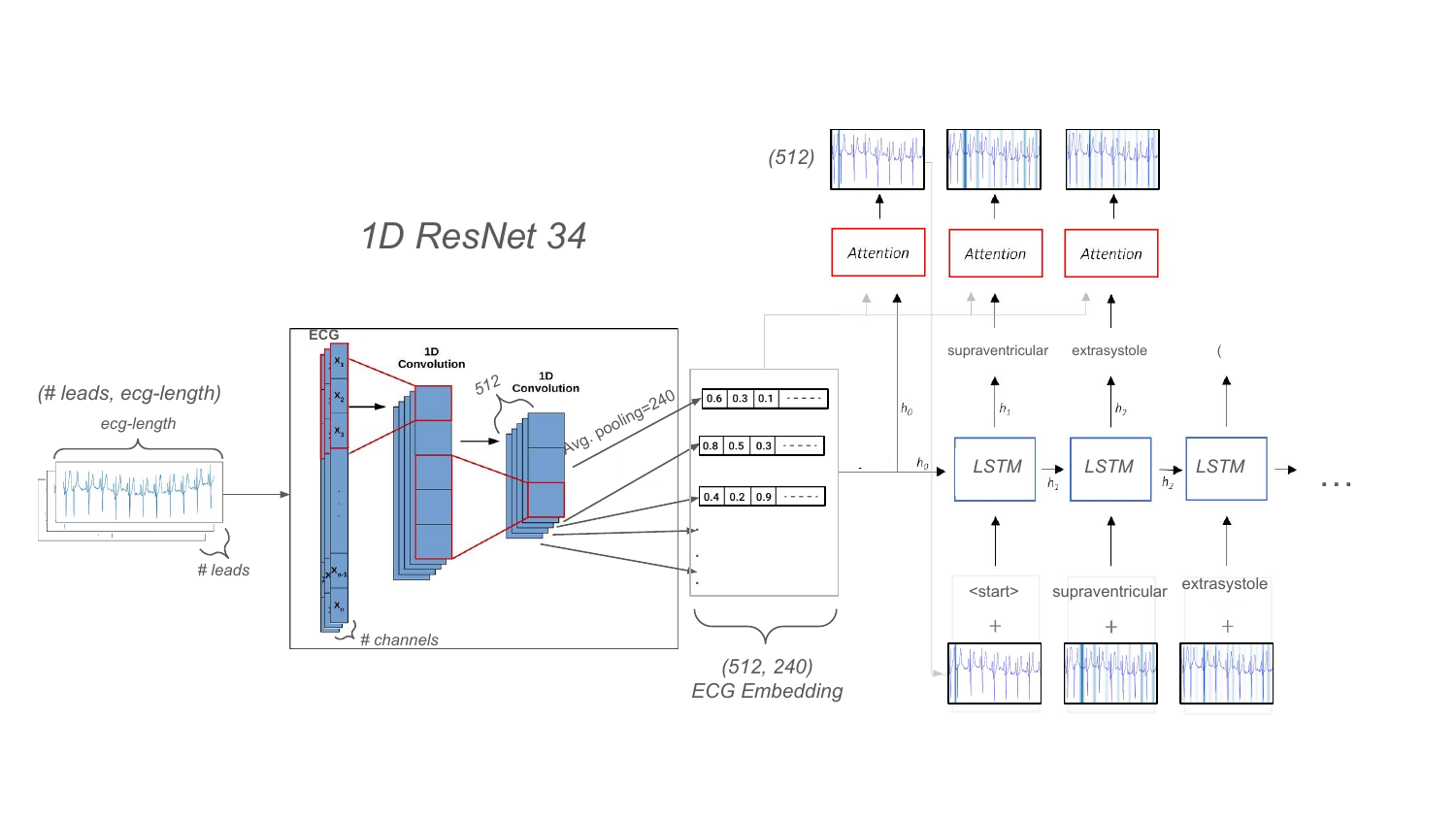}
        \caption{Our LSTM-based model architecture: The ECG signal is input into the 1D ResNet34, producing 512 channels, each of a user-defined size $x$ ($x=240$ in our best-performing model). An attention network aggregates the $x$ temporal data points into a single value per channel, forming a 512-length embedding vector. The vector and current network state are used for predicting the next word. \textit{Note: The ECG episode visualized at each decoding step demonstrates the attention-weighted aggregation (weights are proportional to the intensity of the blue highlight overlay). In practice, the ECG is encoded at this point.}}
        \label{fig:lstm_architecture}
    \end{subfigure}
    
    \caption{Architectures of our (a) Transformer-based and (b) LSTM-based models for ECG signal analysis.}
    \label{fig:model_architectures}
\end{figure*}

The entire encoder-decoder architecture is visualized in Fig. \ref{fig:model_architectures}.

\subsection{Encoder for ECG Embedding} \label{sec:encoder}

To create an effective embedding for ECG signals, we employed a modified 34-layer ResNet architecture tailored for 1D inputs. The standard classification layer was removed and replaced with an average pooling layer, allowing for adjustable output sizes to preserve temporal information across 512 output channels. This adaptation enables the retention of temporal relationships, which together with attention mechanism in the decoder, can be utilized for ECG interpretation. The output embedding from this encoder forms the input to the decoding stage.

\subsection{Decoder Architectures for Generating ECG Description} \label{sec:decoder}

The decoder's task is to generate meaningful, contextual descriptions from the ECG embeddings. We tested two architectures for this purpose: a Transformer-based decoder and an LSTM-based decoder, described in detail below.

\subsubsection{\textbf{Transformer Decoder}}
The Transformer decoder \cite{NIPS2017_3f5ee243} receives a combined input of ECG embeddings and tokenized report sequences, leveraging self-attention to dynamically focus on different parts of the ECG signal during token prediction. The concatenation is formalized as follows:
\begin{equation} 
    \mathbf{X} = [\mathbf{f}_1, \mathbf{f}_2, \dots, \mathbf{f}_K, \mathbf{e}_1, \mathbf{e}_2, \dots, \mathbf{e}_T] \label{eq:transformer_input} 
\end{equation} 
where $\mathbf{e}_t$ denotes the embedding of the $t$-th token in the report, and $\mathbf{f}_k$ represents the features of the $k$-th ECG segment provided by the encoder. $K$ is set by the user and determines the extent of downsampling of the ECG signal. 

Using the combined input $\mathbf{X}$, the transformer calculates the context vector $\mathbf{c}_t$ at each time step $t$:
\begin{equation}
\mathbf{c}_t = \sum_{i=1}^{K+T} \alpha_{t, i} \mathbf{x}_i
\label{eq:transformer_context}
\end{equation}
Where $\mathbf{x}_i$ represents each element in $\mathbf{X}$ which can be either an ECG-segment feature vector $\mathbf{f}_k$ or a token embedding $\mathbf{e}_t$. The attention weights $\alpha_{t,i}$ are calculated following the standard Transformer attention mechanism described in \cite{NIPS2017_3f5ee243}.

\subsubsection{\textbf{LSTM Decoder}} The LSTM decoder \cite{hochreiter1997long} generates one word at a time, conditioned on a context vector, the previous LSTM hidden state, and the last predicted token (or ground truth token during training). The attention mechanism dynamically weights different segments of the encoded ECG signal, creating a context vector $\mathbf{c}_t$ at each time step $t$: 
\begin{equation}
\mathbf{c}_t = \sum_{k=1}^{K} \alpha_{t, k} \mathbf{f}_k
\label{eq:lstm_context}
\end{equation}
where $\alpha_{t,k}$ is the attention weight at time step $t$ for temporal segment $k$, defined by:
\begin{equation}
\alpha_{t, k} = \frac{\exp(\mathbf{w}^T (\mathbf{W}_1 \mathbf{h}_{t-1} + \mathbf{W}_2 \mathbf{f}_k)^+)}{\sum_{j=1}^{K} \exp(\mathbf{w}^T (\mathbf{W}_1 \mathbf{h}_{t-1} + \mathbf{W}_2 \mathbf{f}_j)^+)}
\label{eq:alpha}
\end{equation}
Here, $\mathbf{w}$, $\mathbf{W_1}$ and $\mathbf{W_2}$ are learnable weights within linear layers, and $\mathbf{h}_{t-1}$ is the previous hidden state. 

Ultimately, both the transformer and LSTM based models are trained using cross-entropy loss.

\noindent \textbf{Attention:}\label{par:attention} 
In the realm of natural language processing (NLP), attention mechanisms have been shown to enhance the recognition of objects in images, as demonstrated by Ba et al. \cite{ba2014multiple}. These mechanisms, which focus on relevant parts of the data, are hypothesized to enable the model to dynamically focus on relevant portions of the ECG signal and thus improve the quality of generated descriptions.

While self-attention is natively integrated into the Transformer’s structure and can be adjusted to attend to different segments of the ECG (refer to equation \ref{eq:transformer_context}), it is externally added in our LSTM-based model. In this case, in each prediction step $t$ we use an $\alpha_t$ vector, sized to match the size of each channel output by the encoder (240 in our best-performing model, which represents temporal encoding of the ECG signal), and perform weighted aggregation of it. This results in a single weighted value for each encoder output channel, conditioned on the networks previous state $h_{t-1}$. Refer to equation \ref{eq:alpha} for details about the calculation of $\alpha_t$. By applying weighted aggregation to the channels of the downsampled ECG signal using weights that account for both the ECG signal and the current network state, the model is able to attend to different parts of the ECG when predicting each word in the generated report. An example to this attention driven prediction process is visualized in Fig. \ref{fig:method_overview}. Finally, as in \cite{xu2015show}, doubly stochastic attention regularization is applied to prevent overfitting. This encourages the model to distribute its attention more uniformly, avoiding a tendency to over-focus on specific time steps and ensuring diverse attention across the signal.

\subsection{Training Optimization and Enhancement Techniques}

Our training pipeline incorporates several optimization strategies to improve model performance and ensure reliable generation of ECG descriptions. Key techniques include encoder pre-training and targeted hyperparameter tuning.

\vspace{0.5cm}

\noindent \textbf{Encoder Pre-Training:}
In one configuration, we pre-trained the encoder using rhythm class labels as a preliminary task to strengthen ECG embeddings before fine-tuning on free-text reports. This pre-training phase was intended to leverage rhythm labels for additional contextual learning; Moreover, as described in Section \ref{sec:data} the ICM dataset includes approximately 738K episodes with rhythm class labels but no accompanying reports, enabling us to tap into this large dataset, which would otherwise not be utilized. Pre-training resulted in an F-score of 0.69 on the PTB-XL dataset (6 classes) and 0.49 on the ICM dataset (11 classes).

\vspace{0.5cm}

\noindent \textbf{Hyperparameters Tuning:}
Extensive hyperparameter tuning was conducted to optimize model performance and stability. Key parameters were adjusted as follows:

\begin{itemize}
    \item \textbf{Encoder architecture}: We tested both ResNet-18 and ResNet-34, aiming to optimize both model performance and computational efficiency. 
    \item \textbf{Decoder architecture}: We varied the layer depth for both LSTM and Transformer decoders, from a single layer up to 20 layers.
    \item \textbf{Learning rates}: Separate learning rates were optimized for the encoder and decoder. 
    \item \textbf{Stopping Criteria}: We explored various early stopping criteria such as cross-entropy loss (which was also used for back-propagation), BLEU-1, BLEU-4 and METEOR scores.
    \item \textbf{Additional Parameters:} We also tuned various parameters, including embedding sizes, batch sizes, teacher-forcing probabilities \cite{williams1989learning}, normalization techniques (at batch, dataset, and episode levels), and top-k sampling. These adjustments contributed to enhancing model robustness and performance consistency.
\end{itemize}
\noindent
This comprehensive tuning process involved multiple repetitions of experiments to reduce the impact of random variations, resulting in a stable and reliable evaluation of model performance. The best-performing parameter configurations are as follows:

\vspace{0.3cm} 

\noindent \textbf{Encoder:}
A learning rate of $4 * 10^{-4}$ was optimal, with a stem kernel of 9 in the first convolutional layer and subsequent ResNet stage kernel sizes of 9, 7, 7, and 5, respectively. Each of the 512 output channels was set to an output size of 240.

\noindent 
\textbf{Transformer:}
The best configuration used 12 layers and 8 attention heads, a learning rate of $1 * 10^{-4}$, and an ECG input size of $K=1$ (as described in Section \ref{sec:decoder}, subsection \textit{Transformer Decoder}).

\noindent 
\textbf{LSTM:}
We used a single LSTM layer with a learning rate of $4 * 10^{-4},$ a teacher forcing probability of 1 during training (disabled during validation and testing), and an attention layer of size 512.

\vspace{0.3cm}
\noindent
For both our LSTM and Transformer based models we used batch size of 32, no normalization, cross-entropy as the loss function and Adam optimizer. The token embedding size was set to 512, the dropout rate to 0.5 and the learning rate decay factor to 0.8 every 8 epochs without improvement, with the METEOR score (see section \ref{sec:metrics} under METEOR) as the target metric. Top-k sampling of 1 was used for deterministic token selection.

\subsection{Complexity \& Runtime}
The model was trained on a single node equipped with 8 CPU cores, 62.5 GB of memory, and one NVIDIA A100 GPU (80 GB memory). The runtime was estimated based on training both the encoder and decoder from scratch (without pre-training) on the PTB-XL dataset, using the optimal hyperparameters described in section \ref{sec:method}.

\noindent \textbf{Encoder:} The ResNet-34 architecture contains 13.76M parameters.

\noindent \textbf{Decoder:}
The Transformer decoder, with 38.36M parameters, has an estimated runtime of approximately 7 hours and 10 minutes on the specified hardware for training the entire model, including the encoder (without encoder pre-training). By comparison, the LSTM decoder—with 5.52M parameters—completes training in roughly 1 hour and 10 minutes (in both cases, runtime depends on the number of epochs before early stopping).

While the best-performing Transformer model uses 12 layers, reducing this to 8 layers provides a runtime-optimized alternative with minimal impact on performance. This modification reduces the parameter count to 25.75M and shortens the runtime accordingly. Similarly, using ResNet-18 instead of ResNet-34 reduces the encoder’s parameter count to 6.93M without substantially affecting accuracy.

Note: Both the Transformer and the LSTM training are not optimized in terms of runtime and, presumably a relatively minor effort could decrease it significantly. 
Furthermore, using a pre-trained encoder reduces the runtime by about 10\% (not including the time for pre-training, which, if included, results in a longer training time overall).

\section{Data \& Preprocessing} 
\label{sec:data_and_preprocessing}

In this section, we introduce the datasets used for training and evaluating our ECG caption generation model. We outline the main characteristics of each dataset, followed by the preprocessing techniques employed to ensure compatibility with our model architecture. These steps are critical for enhancing model performance and ensuring robust, generalizable results.

\subsection{Data}\label{sec:data}

The success of our model relies on access to annotated ECG data that captures a wide range of cardiac conditions and data quality.

To achieve this, we selected two complementary datasets: the publicly available PTB-XL dataset, which provides clinical-grade, multi-lead ECG recordings, and the proprietary dataset provided by BIOTRONIK SE \& Co KG, comprising single-lead sECG data from implantable cardiac monitor (ICM) devices. This combination allows us to evaluate the model's adaptability to varying ECG formats, sampling rates, and recording environments, thus ensuring its relevance for both clinical and personal health monitoring applications.

\subsubsection{\textbf{PTB-XL}}

The PTB-XL dataset \cite{wagner2020ptb,goldberger2000physiobank} comprises 21,801 10-second ECG episodes, with two optional sampling frequencies: 500 Hz (resulting in 5000 data points per ECG record) and 100 Hz (1000 data points per record). These episodes were collected from 18,869 unique patients, giving a \textit{unique patient proportion} of 0.87 - indicating that 87\% of the total episode count corresponds to unique individuals, while the remaining episodes are contributed by patients with multiple recordings. Each episode is annotated with a rhythm label and a corresponding report (for details see \cite{wagner2020ptb}).

The dataset includes 9839 unique reports, where uniqueness is determined by ignoring minor differences such as whitespace, case sensitivity, and punctuation. This results in a unique report proportion of 0.45, meaning that 45\% of the total reports are distinct, with the remainder being repeated across multiple episodes. Using this dataset served multiple purposes in our research. Firstly, it enabled us to test our model on a publicly available dataset, ensuring the reproducibility of our results. Additionally, it allowed us to assess the model's performance on 12-lead, 500 or 100 Hz, 10-second ECG records and evaluate how it manages a relatively small sample size. Furthermore, we used this dataset for comparability with the reference model, which was also tested on PTB-XL \cite{qiu-etal-2023-transfer} (on 100Hz).

\subsubsection{\textbf{Implantable Cardiac Monitor (ICM)}}

The ICM dataset contains records from Implantable Cardiac Monitors (ICMs) BIOMONITOR III and BIOMONITOR IIIm \cite{web:biomonitor}, which consists of 60-second sECG episodes recorded at a sampling frequency of 128 Hz, resulting in 7680 data points per sECG record. The report and rhythm class (a single label selected by HCPs out of given list) data were assembled as a byproduct of routine medical procedures conducted worldwide. The anonymized data underwent additional filtering to ensure that all personal information was removed from the manually entered reports. To avoid translation issues only reports in English were used. The final report dataset (\textbf{Part 2} below) comprised data from 1033 clinics. The dataset is divided into two parts:

\begin{itemize}
    \item \textbf{Part 1:} 737,999 episodes, each labeled with a rhythm class provided by an HCP. These labels are a byproduct of routine medical procedures and were not cross-checked for accuracy and are therefore potentially inaccurate.
    \item \textbf{Part 2:} 206,768 episodes, where each episode includes both a rhythm class and a free-text report: a sentence written by an HCP as part of routine medical procedures, not necessarily intended as an ECG caption for learning or training purposes. These episodes come from 6687 implanted devices. Among these 200k episodes, there are 38,652 unique reports, (here too, ignoring minor differences like whitespace, case sensitivity, and punctuation), resulting in a unique report proportion of 0.19.
\end{itemize}

\subsection{Data Pre-Processing}
\label{sec:pre-processing}

Consistent data preprocessing ensures compatibility with our model architecture and optimizes performance by standardizing data across both datasets. Steps such as train-validation-test splits, abbreviation unification, and tokenization were carefully designed to maintain data integrity while enhancing comparability.

\paragraph{Episode Deduplication:}
In the ICM dataset, we applied an episode deduplication process. An episode was considered duplicated if it originated from the same implanted device, had the same report, rhythm class, and recording date. During deduplication, one episode from each group of duplicates was randomly selected, and the others were removed from the dataset.

\paragraph{Train/Validation/Test Splits:}

To prevent data leakage and enhance model generalizability, we applied device-level splits for training, validation, and testing. This approach ensures that each patient’s data is confined to one split, reducing the risk of overfitting and supporting a more meaningful evaluation of the model’s performance.

\begin{itemize}
    \item \textbf{PTB-XL:} We tested two splitting approaches for this dataset. The first approach used the official splits provided by PhysioNet, which maintain patient exclusivity across splits and mitigate data leakage risks. These splits allocate 80\% of data for training, 10\% for validation, and 10\% for testing, as further detailed in \cite{wagner2020ptb}. The second approach, used for comparability with the reference model \cite{qiu-etal-2023-transfer}, involved a random split into training, validation, and testing sets with proportions of 64\%, 16\%, and 20\%, respectively. Results from both approaches are presented in Section \ref{sec:results}, Tables \ref{table:results-ptb-xl_official} and \ref{table:results-baseline_compare}.

    \item \textbf{ICM:} For this dataset, splits were applied at the implanted device level to ensure no overlap between the training, validation, and test sets, preventing the model from leveraging morphological similarities across episodes from the same device. After deduplication, data was divided into training (80\%), validation (10\%), and test (10\%) sets.
\end{itemize}

\vspace{0.3cm}
\paragraph{Abbreviations \& Translations}
Due to medical terminology variations in the datasets, an abbreviation unification process was applied to reduce prediction errors. Additionally, the OPUS-MT model \cite{tiedemann2023democratizing, TiedemannThottingal:EAMT2020} was used to translate German reports from the PTB-XL dataset into English, allowing for consistent application of abbreviation standardization and enhancing model compatibility. The abbreviations and their unified form are presented in Table \ref{table:abbreviations}.

\begin{table}[h]
    \centering
    \begin{tabular}{|l|l|}
        \hline
        \textbf{Replaced Term} & \textbf{Unified Form} \\ \hline
        @                     & at                    \\ \hline
        +, \&                 & and                   \\ \hline
        pac, pacs, sve, sves, apc, apcs & premature atrial contraction \\ \hline
        pvc, pvcs, vpc, vpcs, ves & premature ventricular contraction \\ \hline
        ectopic, ectopics, ectopy & premature contraction  \\ \hline
        brady                 & bradycardia            \\ \hline
        sb                    & sinus bradycardia      \\ \hline
        tachy, tachycardia   & tachycardia            \\ \hline
        st                    & sinus tachycardia      \\ \hline
        svt                   & supraventricular tachycardia \\ \hline
        nsvt                  & nonsustained ventricular tachycardia \\ \hline
        sr                    & sinus rhythm           \\ \hline
        nsr                   & normal sinus rhythm    \\ \hline
        af, afib, a-fib      & atrial fibrillation    \\ \hline
        afl, a-flutter, aflutter & atrial flutter       \\ \hline
        cw                    & continuous wave        \\ \hline
        rvr                   & rapid ventricular rate \\ \hline
        ppm                   & permanent pacemaker    \\ \hline
        bpm                   & beats per minute       \\ \hline
        pat, pt               & patient                \\ \hline
        bts                   & beats                  \\ \hline
        wo, w/o               & without                \\ \hline
        w, w/                & with                   \\ \hline
        hr                    & heart rate             \\ \hline
        avb                   & atrioventricular block \\ \hline
        \hline
    \end{tabular}
    \caption{List of terms/abbreviations and their unified forms, as used for the abbreviation unification process described in Section \ref{sec:data} under Abbreviations \& Translations. Terms are presented in lowercase, but both upper and lower case variants are treated equivalently.}
    \label{table:abbreviations}
\end{table}

\subsection{Report Tokenization}

Report tokenization was conducted by splitting comments based on non-letter characters (such as spaces, commas, slashes, etc.). These splitting characters were tokenized and retained within the text, with the exception of spaces. Spaces were not tokenized but assumed to occur between every two tokens to avoid skewing evaluation metrics (e.g., METEOR or BLEU-score) due to the abundance of spaces, which would otherwise inflate scores and reduce the relative significance of other tokens.

This process yielded an initial vocabulary size of 5,304 for the ICM dataset and 2,282 for the translated PTB-XL dataset, or 7,015 and 2,455, respectively, without applying the abbreviation script. As abbreviations only apply to English, they were not used on the non-translated PTB-XL reports. Filtering out words occurring fewer than twice further reduced the ICM and PTB-XL datasets vocabularies to 3,194 and 1,383 tokens, respectively. To ensure comparability across datasets, we truncated each vocabulary, retaining only the 1,024 most frequent tokens. Note that after vocabulary truncation, the least frequent word of the PTB-XL has a frequency of 3 and that of the ICM dataset a frequency of 14. Therefore, in the presented case filtering tokens based on their frequency deemed redundant.

Each report was prefixed with a start token and suffixed with an end token, with padding tokens added to a maximum report length of 300 tokens (although in practice, no reports exceeded this length). A special token was assigned for unknown tokens: words appearing only once or outside the top 1,024 most common tokens.

The Byte-Pair-Encoding (BPE) method \cite{sennrich2015neural} was also tested, but it performed suboptimally compared to word tokenization, likely due to the specialized terminology in these datasets.

\vspace{0.3cm}
In summary, the PTB-XL and ICM datasets provide diverse ECG data well-suited for training and evaluation of our caption generation model. By carefully preprocessing the data, including patient/device-level splits, translation and abbreviation unification, we enhanced our model's ability to generate precise, contextually relevant ECG descriptions.

\section{Experiments} \label{sec:experiments}
In this section, we summarize extensive evaluations of our model across multiple design configurations (Section \ref{sec:key_experiments}) and its comparison with a state-of-the-art method (see Section \ref{sec:baseline}). Additionally, we present the results of our case study (Section \ref{sec:exp_biotronik}) and a sanity check conducted to confirm that the model's performance is attributable to its ability to learn ECG morphology (Section \ref{sec:sanity-check}).

\subsection{Key Experiments} \label{sec:key_experiments}
Throughout our experimentation, we addressed several key research questions by running a series of targeted experiments. Table \ref{table:results-ptb-xl_official}, summarizes the performance of our model on the publicly available PTB-XL dataset across the following main variations:
\begin{itemize}
    \item \textbf{Decoder Architecture (LSTM vs. Transformer):} As the choice of decoder architecture is fundamental to our model, we tested all configurations with both LSTM and Transformer decoders, allowing us to compare overall performance between the two.
    \item \textbf{Encoder Depth (ResNet18 vs. ResNet34):} To determine whether the depth of the ResNet encoder affects model performance, we experimented with both ResNet18 and ResNet34 encoders.
    \item \textbf{Effect of Encoder Pre-Training:} To quantitatively evaluate whether pre-training the encoder on rhythm class labels improves model performance, we compared results with and without encoder pre-training. In our case-study, as discussed in Section \ref{sec:data}, such pre-training also leverages an additional 738K episodes.
    \item \textbf{Translation to English:} Since our study primarily uses English as the target language, we conducted all experiments with text translated to English. To maintain reproducibility, however, we also ran an experiment without translation or abbreviation unification, testing the model on raw text.
    \item \textbf{Abbreviation Unification vs. Raw Text:} To test the impact of the abbreviation unification process described in Section \ref{sec:pre-processing}, we compared results on such standardized text against results on the raw text. As the abbreviation unification applies exclusively to English, we included translation for this comparison as well.
\end{itemize}

\subsection{Comparison with the Reference Model} 
\label{sec:baseline}
To evaluate the effectiveness of our approach, we benchmarked its performance against the reference model proposed by Qiu et al. \cite{qiu-etal-2023-transfer}. This model employs a ResNet-based encoder for feature extraction from ECG signals, which are then aligned with pretrained embeddings from large language models (LLMs) such as GPT-2 and BERT. Additionally, Qiu et al. introduced an Optimal Transport (OT)-based objective complementing the standard cross-entropy loss, to enhance alignment between ECG embeddings and language embeddings. Their method has achieved strong results in ECG disease classification and report generation tasks, particularly on the PTB-XL dataset, making it a relevant benchmark for comparison.

However, Qiu et al.’s use of random train/validation/test splits raises concerns regarding patient data overlap, potentially inflating results due to shared patient data across splits. Such overlap is problematic for generalizability, a critical factor for clinical applications. To address this, we adopt more rigorous splitting methods in our experiments, including the official PTB-XL splits that ensure patient exclusivity across sets, thereby enhancing reliability and comparability.

A detailed comparison between the reference model and our model’s performance is provided in Section \ref{sec:results}, particularly in Table \ref{table:results-baseline_compare}.

\subsection{Experimental Setup} \label{sec:exp-setup}
Our primary experiments were conducted on the PTB-XL dataset, which serves as a comprehensive benchmark in this domain. To ensure reproducibility and generalizability, we used the train/validation/test splits recommended by the PTB-XL authors. These fixed splits enhance reliability as they are not random, and the validation and test datasets are considered gold standards, comprising reports manually reviewed by professionals \cite{wagner2020ptb}.

To facilitate a direct comparison with the reference model, which is evaluated on random splits, we also tested our best-performing models on comparable random splits, applying translation but excluding abbreviation unification to align with the reference approach. Results from this experiment are presented in Table \ref{table:results-baseline_compare}.

Additionally, as a case-study, we evaluated our model on the ICM dataset. This dataset represents an increasingly relevant data source generated by personal ECG devices, and its automated analysis is of growing importance \cite{bleich2023enhancing}. The results for this case-study can be found in Table \ref{table:results-biotronik}.

For detailed explanation of the different datasets used in the different setups please refer to section \ref{sec:data_and_preprocessing}.

Lastly, we performed a \textit{sanity check} by substituting the ECG input with a uniform vector in which all entries are set to 1, effectively removing the ECG signal to assess whether the model was generating reports based on repetitive textual patterns rather than meaningful ECG data. The performance of our model in this test is discussed in Section \ref{sec:sanity-check}.

\subsection{Evaluation} \label{sec:metrics}

Evaluating generated text is a challenging task, as even humans may struggle to assess the similarity in meaning between two sentences. We used three key metrics suited for such evaluation:

\textbf{BLEU} \cite{papineni2002bleu} measures the precision of N-grams (1- to 4-grams) between generated and reference texts, applying a brevity penalty for shorter outputs to account for recall. This metric is widely used for its simplicity, but it emphasizes precision over recall.

\textbf{METEOR} \label{par:meteor} \cite{banerjee2005meteor} improves on BLEU by incorporating both precision and recall, offering a more balanced assessment. It also includes word-order sensitivity and explicit word-matching (exact, stemmed, and synonym-based), making it suitable for tasks where accurate phrasing is critical.

\textbf{ROUGE} \cite{lin2004rouge}, originally designed to emphasize recall (R), is particularly useful for assessing the completeness of information, as is often required in summarization tasks. However, beyond Recall, ROUGE also includes Precision (P) and F1-score (F), providing a more comprehensive evaluation of generated text.

Another metric worth noting is \textbf{MRScore} \cite{liu2024mrscore} (preprint), which utilizes large language models (LLMs) such as GPT to evaluate generated reports based on human-like criteria. While it has not yet gained widespread adoption in clinical text generation, it represents a promising direction for incorporating LLM-based assessments in the future.

METEOR was chosen as our primary metric due to its ability to evaluate semantic similarity more effectively than BLEU or ROUGE. BLEU's reliance on exact matches makes it less suitable for tasks like clinical text generation, where paraphrasing or synonym usage is common. ROUGE, while including precision and F1-score, is primarily recall-focused, which can bias evaluations towards longer outputs. METEOR, on the other hand, balances precision and recall while incorporating linguistic features like stemming and synonyms, making it better suited for capturing subtle nuances in clinical reports.

\section{Results}\label{sec:results}


\begin{table*}[ht]
    \centering
    \caption{Performance results for key experiments on the PTB-XL dataset (official splits)}
    \begin{tabularx}{\textwidth}{l|l|l|>{\centering\arraybackslash}X|>{\centering\arraybackslash}X>{\centering\arraybackslash}X>{\centering\arraybackslash}X|>{\centering\arraybackslash}X>{\centering\arraybackslash}X>{\centering\arraybackslash}X}
        \toprule
        \multirow{2}{*}{Encoder} & \multirow{2}{*}{Decoder} & \multirow{2}{*}{Additional} & \multirow{2}{*}{METEOR} &  \multicolumn{3}{c|}{BLEU (\%)} & \multicolumn{3}{c}{ROUGE-1 (\%)} \\
        \cmidrule(lr){5-7} 
        \cmidrule(lr){8-10} 
        & & & (\%) & 1 & 2 & 4 & P & R & F \\
        \midrule
        ResNet34 & LSTM & Trans./Abbr. & \textbf{55.53} & \textbf{51.63} & \textbf{44.54} & 35.29 & 61.47 & \textbf{60.65} & 58.33 \\
        ResNet34 & LSTM & Trans. & 55.01 & 51.03 & 43.72 & 34.39 & 61.76 & 59.65 & 57.93 \\
        ResNet34 & LSTM & - & 50.72 & 48.25 & 42.29 & 34.90 & 55.53 & 54.93 & 52.29 \\
        ResNet34 & Transformer & Trans./Abbr. & 55.00 & 51.41 & 44.04 & 34.62 & 61.74 & 59.82 & 58.12 \\
        ResNet34 & Transformer & Trans. & 54.84 & 50.00 & 42.76 & 33.64 & 62.74 & 59.04 & 58.18 \\
        ResNet34 & Transformer & - & 51.11 & 47.21 & 41.02 & 33.29 & 57.61 & 54.76 & 53.87 \\
        Pre. ResNet34 & LSTM & Trans./Abbr. & 53.19 & 48.75 & 41.46 & 32.26 & 59.54 & 57.90 & 55.82 \\
        Pre. ResNet34 & Transformer & Trans./Abbr. & 52.83 & 48.93 & 41.42 & 31.93 & 60.77 & 57.96 & 56.73 \\
        ResNet18 & LSTM & Trans./Abbr. & 55.30 & 51.40 & 44.39 & \textbf{35.37} & 62.40 & 59.57 & 58.33 \\
        ResNet18 & Transformer & Trans./Abbr. & 55.00 & 50.95 & 43.70 & 34.53 & \textbf{63.47} & 59.39 & \textbf{58.82} \\ \midrule
        \multicolumn{2}{c|}{} Reference (Qiu et al.)* &  & 24.51 & 27.21 & - & - & 26.12 & 35.71 & 29.56 \\
        \bottomrule
    \end{tabularx}
    \caption*{Results from key experiments (Section \ref{sec:key_experiments}) on the PTB-XL dataset using official PhysioNet-suggested splits \cite{wagner2020ptb}. Bold values denote the highest scores for each metric across experiments. "Pre." indicates a pre-trained encoder. "Trans." denotes translated reports, and "Abbr." denotes abbreviation unification.
    *Note: Qiu et al. evaluated their model performance on random splits, using a Transformer encoder and BART decoder. For comparability, our model’s results on random splits are in Table \ref{table:results-baseline_compare}.}
    \label{table:results-ptb-xl_official}
\end{table*}

\subsection{Key experiments}
The results of our main experiments, as shown in Table \ref{table:results-ptb-xl_official}, indicate that our model outperforms the current state-of-the-art reference method across all metrics. This improvement is achieved even under a more rigorous setup, using patient-exclusive test sets, compared to the random splits used in the reference model.

Our findings suggest that the optimal configuration employs a non-pretrained ResNet34 as the encoder and an LSTM as the decoder. However, performance across other configurations did not differ significantly. Notably, the performance gap between ResNet34 and the more compact ResNet18 encoder was minimal, suggesting that ResNet18 is a practical alternative when hardware or time constraints are considerations. This setup, combined with an LSTM decoder and bypassing time-intensive translation steps, still yielded comparable results.

Moreover, pre-training the encoder on episode rhythm labels did not improve model performance; in fact, it appeared to slightly diminish predictive accuracy. Translating the reports to English, however, resulted in a noticeable METEOR score boost from 50.72\% for the non-translated version to 55.01\% for the translated version in the top-performing architecture. Abbreviation unification had a minor yet positive impact (about 0.5\%) on the scores for translated reports.

\begin{figure}[H]
    \centering
    \caption*{\textbf{METEOR Score Progression for Main Experiments on PTB-XL (Official Splits)}}
    \vspace{0.1cm}
    \includegraphics[width=\columnwidth]{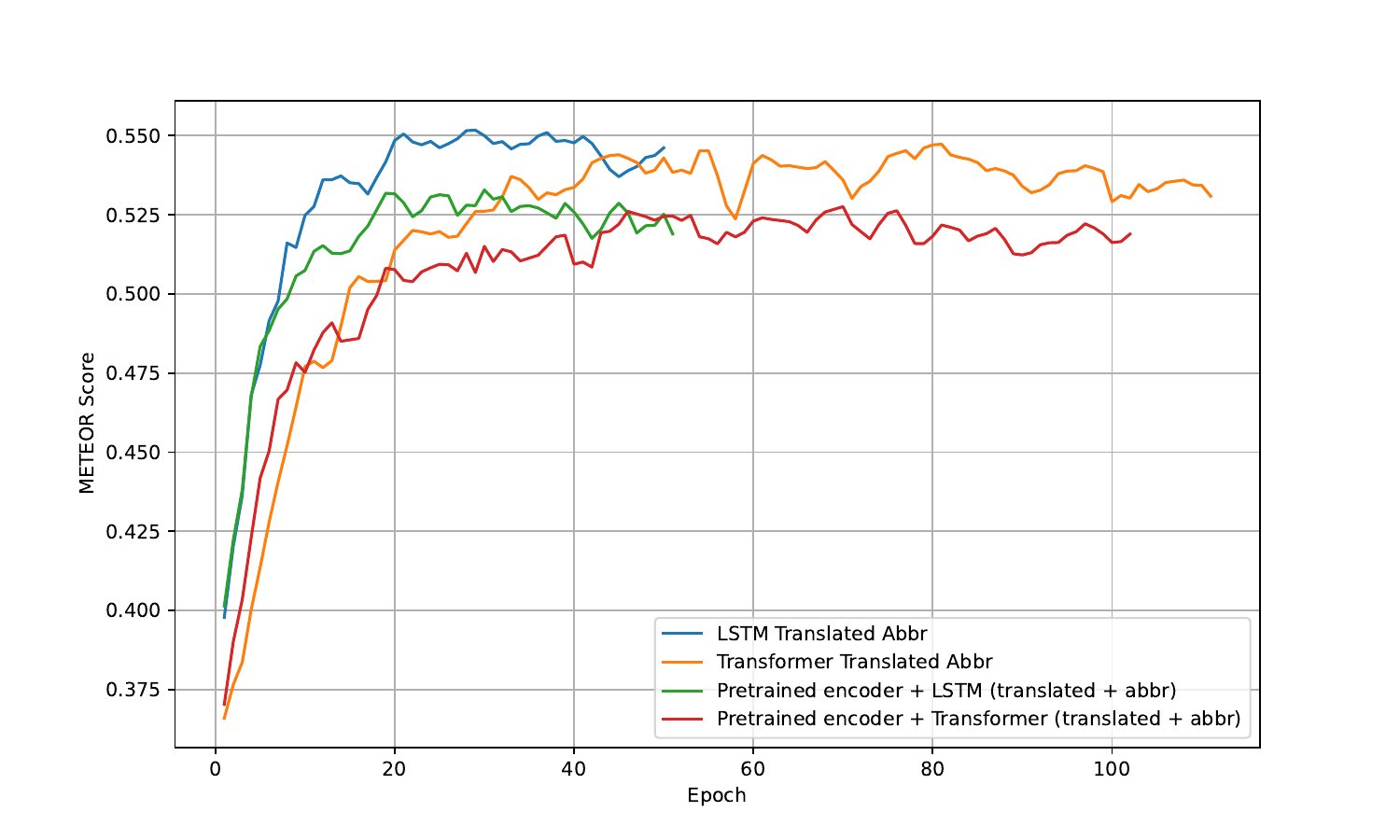}
    \caption{Progression of validation METEOR scores across training epochs on official validation split of the PTB-XL dataset, over several key experiments (Abbr. stands for abbreviation unification - see Section \ref{sec:pre-processing} in paragraph Abbreviations \& Translations for details). Scores are smoothed using a moving average with a window size of 3 for clarity. Early stopping was implemented at 30 epochs, indicating that peak performance was reached 30 epochs prior to the line endpoints. Exact METEOR scores from these experiments are provided in Table \ref{table:results-ptb-xl_official}.}
    \label{fig:ptbxl_main_exp_meteor}
\end{figure}

Fig. \ref{fig:ptbxl_main_exp_meteor} shows the progression of the METEOR score over training epochs for key experiments. The plot suggests the absence of overfitting and reveals that the LSTM-based model achieves accuracy comparable to the Transformer-based model in nearly half the epochs.

\begin{table}[ht]
    \centering
    \caption{Results of reference method vs our method on PTB-XL - random splits}
    \begin{tabularx}{\columnwidth}{l|c|c|ccc}
        \toprule
        \multirow{2}{*}{Experiment} & \multirow{2}{*}{METEOR} & \multirow{2}{*}{BLEU-1} & \multicolumn{3}{c}{ROUGE-1 (\%)} \\
        \cmidrule(lr){4-6} 
        & (\%) & (\%) & P & R & F \\
        \midrule
        Reference (Qiu et al.) & 24.51 & 27.21 & 26.12 & 35.71 & 29.56 \\
        ResNet34 + LSTM & \textbf{63.01} & \textbf{58.84} & \textbf{68.94} & \textbf{67.24} & \textbf{65.82} \\
        ResNet34 + Tran. & 62.00 & 57.54 & 68.09 & 66.22 & 64.88 \\
        \bottomrule
    \end{tabularx}
    \caption*{Results of the reference method compared with our best-performing LSTM and Transformer (denoted as "Tran.") architectures on the PTB-XL dataset, evaluated using random train/validation/test splits. Bold values highlight the highest metrics achieved among the experiments. Preprocessing steps, including report translation without abbreviation unification (refer to Section \ref{sec:pre-processing}), were standardized across experiments for comparability. Our models, particularly the LSTM-based architecture, consistently outperformed the reference model across all evaluated metrics.}
    \label{table:results-baseline_compare}
\end{table}

\subsection{Experiment with the Reference Model Setup}
To ensure a fair comparison, we conducted an additional experiment replicating the setup used by the reference model (detailed in Section \ref{sec:exp-setup}). The results, presented in Table \ref{table:results-baseline_compare}, demonstrate that both our LSTM-based and Transformer-based models outperform the reference model, with the LSTM-based model showing a slight advantage over the Transformer-based approach.

\begin{table*}[ht]
    \centering
    \caption{Results of our method on the ICM dataset}
    \begin{tabularx}{\textwidth}{l|l|c|>{\centering\arraybackslash}X|>{\centering\arraybackslash}X>{\centering\arraybackslash}X>{\centering\arraybackslash}X|>{\centering\arraybackslash}X>{\centering\arraybackslash}X>{\centering\arraybackslash}X}
        \toprule
        \multirow{2}{*}{\shortstack{Encoder}} & \multirow{2}{*}{\shortstack{Decoder}} & \multirow{2}{*}{\shortstack{Abbr.}} & \multirow{2}{*}{METEOR} & \multicolumn{3}{c|}{BLEU (\%)} & \multicolumn{3}{c}{ROUGE-1 (\%)} \\
        \cmidrule(lr){5-7} 
        \cmidrule(lr){8-10} 
        & & & (\%) & \shortstack{1} & \shortstack{2} & \shortstack{4} & \shortstack{P} & \shortstack{R} & \shortstack{F} \\
        \midrule
        ResNet34 & LSTM & + & 31.90 & 28.50 & 22.86 & 10.63 & \textbf{47.16} & 40.63 & 39.66 \\
        Pre. ResNet34 & LSTM & + & 32.27 & 28.34 & 22.86 & \textbf{10.61} & 47.29 & 40.61 & 39.99 \\
        ResNet34 & Transformer & + & \textbf{32.59} & \textbf{29.00} & \textbf{23.17} & 10.42 & 47.15 & \textbf{41.74} & \textbf{40.22} \\
        Pre. ResNet34 & Transformer & + & 32.20 & 28.68 & 22.96 & 10.56 & 46.49 & 40.94 & 39.66 \\
        ResNet34 & LSTM & - & 15.57 & 11.97 & 5.42 & 1.66 & 31.90 & 27.78 & 26.18 \\
        \bottomrule
    \end{tabularx}
    \caption*{Results of our four model configurations on the ICM test data. "Pre." indicates pre-trained models, and "Abbr." denotes abbreviation unification, as explained in Section \ref{sec:pre-processing} in paragraph Abbreviations \& Translations. Bolded scores mark the highest values across the experiments. The Transformer model without pre-training outperformed the other configurations, with the LSTM model showing competitive performance.}
    \label{table:results-biotronik}
\end{table*}

\subsection{Case Study - Single lead, subcutaneous ECG} \label{sec:exp_biotronik}
In our case-study, we evaluated performance on single-lead, subcutaneous ECG data with reports from ICM devices (See Section \ref{sec:data}). The model’s performance on this dataset, though lower than on PTB-XL, still surpassed current state-of-the-art benchmarks, demonstrating the model's robustness in challenging data scenarios. Notably, the Transformer and LSTM-based models performed similarly, with the Transformer showing a slight edge, likely due to minor random variations.

Notably, despite the 738K episodes containing only rhythm labels and thus available only for pre-training (alongside $\sim$200K episodes with labels as well as reports which are used for both pre-training and training), pre-training the encoder did not enhance performance on this dataset.

The comparatively lower BLEU-4 score for the ICM dataset may be due to the brevity of the reports (median length of 4 tokens and an average of 5.67), reducing 4-gram match probability between generated and reference texts. This is supported by the relatively higher BLEU-1 and BLEU-2 scores, indicating more accurate matching in shorter text segments.

Lastly, abbreviation unification markedly improved performance on this dataset. While its effect was minimal on PTB-XL, it led to a pronounced increase (from 15.57\% to 32.59\%) on the ICM dataset. This is likely because a significant portion (31.2\%) of the PTB-XL dataset is automatically generated, which may result in a higher degree of uniformity in terms and language. In contrast, the ICM dataset exhibits greater variability in terminology, an inconsistency mitigated effectively by abbreviation unification.

\subsection{Sanity Check} \label{sec:sanity-check}
To ensure our models were generating text based on ECG data rather than relying on common text patterns, we performed a sanity check. Here, we replaced the ECG input with a uniform vector in which all entries are set to $1$, effectively removing ECG information. A successful sanity check would result in a substantial drop in performance, confirming that the model's output is informed by ECG morphology. As expected, this experiment led to prediction of uniform comments of optimal length comprising the most common words in the corpus.

The sanity check was conducted both on the PTB-XL and the ICM datasets, post-abbreviation unification. 
The results on the ICM dataset demonstrate an 8.52\% drop in METEOR score, from 32.59\% in the regular experiment to 24.07\% in the sanity check and a drop of 2.11\% in BLEU4 score, from 10.61\% in the regular experiment to 8.5\% in the sanity check, reflecting 26\% and 19\% reduction in these scores respectively. While the relatively high performance in the sanity check suggests some dependency on recurring tokens, the observed drop highlights the model's ability to incorporate meaningful ECG data, even in the relatively challenging ICM dataset.

As expected, the sanity check performed on the official splits of the PTB-XL dataset resulted in a more pronounced performance drop across various metrics. For instance, BLEU4 decreased from 0.35 in the original experiment to 0.08 in the sanity check, reflecting a 77\% reduction. METEOR score dropped from 0.56 to 0.31, a 45\% reduction. 

These results further validate the model's reliance on ECG morphology.


\section{Discussion and Future Work}\label{section:discussion}

The successful application of free-text generation techniques to ECG episodes using an architecture traditionally employed in image captioning represents a notable advancement. It underscores the potential for adapting image-captioning methods to analyze and generate free-text descriptions for medical signals, thereby expanding the scope of cross-domain applications of such techniques.

A central challenge identified in this study is the absence of a well-established, large-scale benchmark dataset specifically for ECG-based report generation. Although the PTB-XL dataset provides a foundation for evaluation, its relatively limited size poses constraints on training more sophisticated language models. An anticipated alternative is the MIMIC-IV-Notes dataset, which is expected to be linked with the MIMIC-IV-ECG dataset \cite{gow2023mimic}. Future evaluation of our proposed model on this dataset could yield a more comprehensive performance assessment, and we anticipate that with larger datasets, the efficiency and accuracy benefits of Transformer-based models may become more pronounced. We also encourage other research groups to use this dataset upon its release to foster a standardized benchmark for the field.

Furthermore, we would like to highlight the importance of our experiments that did not produce our best-performing results, as we believe these acknowledgments are valuable for future research, guiding it toward or away from potential approaches. Examples include our pre-training strategies, translation of reports versus non-translation, lack of abbreviation unification, and the exploration of different combinations thereof. Furthermore, we conducted numerous hyperparameter tests—though not the entire field—due to time constraints and limited computational resources. We encourage future researchers to examine these suboptimal experiments carefully and learn what might or might not be worth attempting in their own studies.

Finally, building on the demonstrated adaptability of image captioning architectures to 1-dimensional medical data in this work, a particularly promising avenue for further exploration lies in applying these methods to other 1D datasets such as EEG (electroencephalogram), as well as data from other domains.

\section{Conclusion}
This work demonstrates the feasibility of generating clinically relevant free-text comments for ECG episodes by employing architectures commonly used in tasks like image captioning. Both LSTM-based and Transformer-based models achieved strong performance, with the LSTM-based model displaying overall advantage due to efficiency and training time.

While these results are promising, they only represent an initial step toward fully automated ECG report generation. The current model, though effective in generating insightful text, is not yet sufficiently refined for standalone use in clinical settings. However, its performance suggests that even in its current state, the model could serve as a valuable assistive tool, providing diagnostic insights to physicians or enabling efficient reporting for large-scale ECG datasets.

In conclusion, this work introduces a state-of-the-art approach to generating ECG reports across a diverse range of episode characteristics, including variations in lead count, resolution, and report quality. The comprehensive pipeline and methodology developed here offer a robust foundation for further research and advancement, supporting future investigations into automated medical report generation and related topics.

\section*{Acknowledgment}

This work was supported by the German Ministry for Education and Research (BMBF) within the Berlin Institute for the Foundations of Learning and Data---BIFOLD (project grants 01IS18025A and 01IS18037I) and the Forschungscampus MODAL (project grant 3FO18501).

\section*{References}

\bibliographystyle{IEEEtran}
\bibliography{main}

\end{document}